\pdfoutput=1

\documentclass[11pt]{article}

\usepackage[preprint]{acl}

\usepackage{times}
\usepackage{latexsym}

\usepackage[T1]{fontenc}

\usepackage[utf8]{inputenc}

\usepackage{microtype}

\usepackage{inconsolata}

\usepackage{graphicx}

\usepackage{tcolorbox}
\usepackage{adjustbox}
\usepackage{booktabs}
\usepackage{verbatim}

\title{End-to-End Aspect-Guided Review Summarization at Scale
}

\author{
\textbf{Ilya Boytsov, \quad Vinny DeGenova, \quad Mikhail Balyasin, \quad Joseph Walt} \\
\textbf{Caitlin Eusden, \quad Marie-Claire Rochat, \quad Margaret Pierson} \\
\\
Wayfair \\
\\
\texttt{\{iboytsov, vdegenova, mbalyasin, jwalt,} \\
\texttt{ceusden, mrochat, mpierson1\}@wayfair.com}
}

\begin{document}
\maketitle
\begin{abstract}
 We present a scalable large language model (LLM)-based system that combines aspect-based sentiment analysis (ABSA) with guided summarization to generate concise and interpretable product review summaries for the Wayfair platform. Our approach first extracts and consolidates aspect–sentiment pairs from individual reviews, selects the most frequent aspects for each product, and samples representative reviews accordingly. These are used to construct structured prompts that guide the LLM to produce summaries grounded in actual customer feedback. We demonstrate the real-world effectiveness of our system through a large-scale online A/B test. Furthermore, we describe our real-time deployment strategy and release a dataset\footnote{https://huggingface.co/collections/IeBoytsov/review-summaries-68dab02e7b6a5bc8e29e81fa} of 11{,}8 million anonymized customer reviews covering 92{,}000 products, including extracted aspects and generated summaries, to support future research in aspect-guided review summarization.
\end{abstract}

\begin{figure}[h!]
\centering
\includegraphics[width=0.5\textwidth]{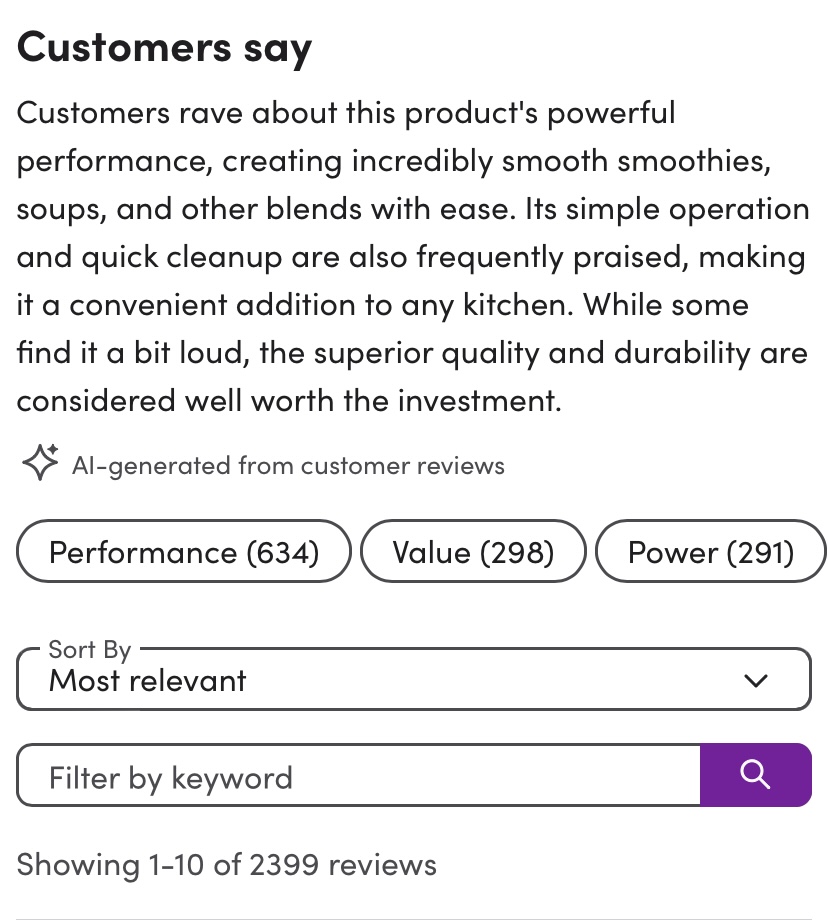} 
\caption{Product Summary example from wayfair.com}
\label{fig:screenshot}
\end{figure}

\begin{figure*}[t]
    \centering
    \includegraphics[width=\textwidth]{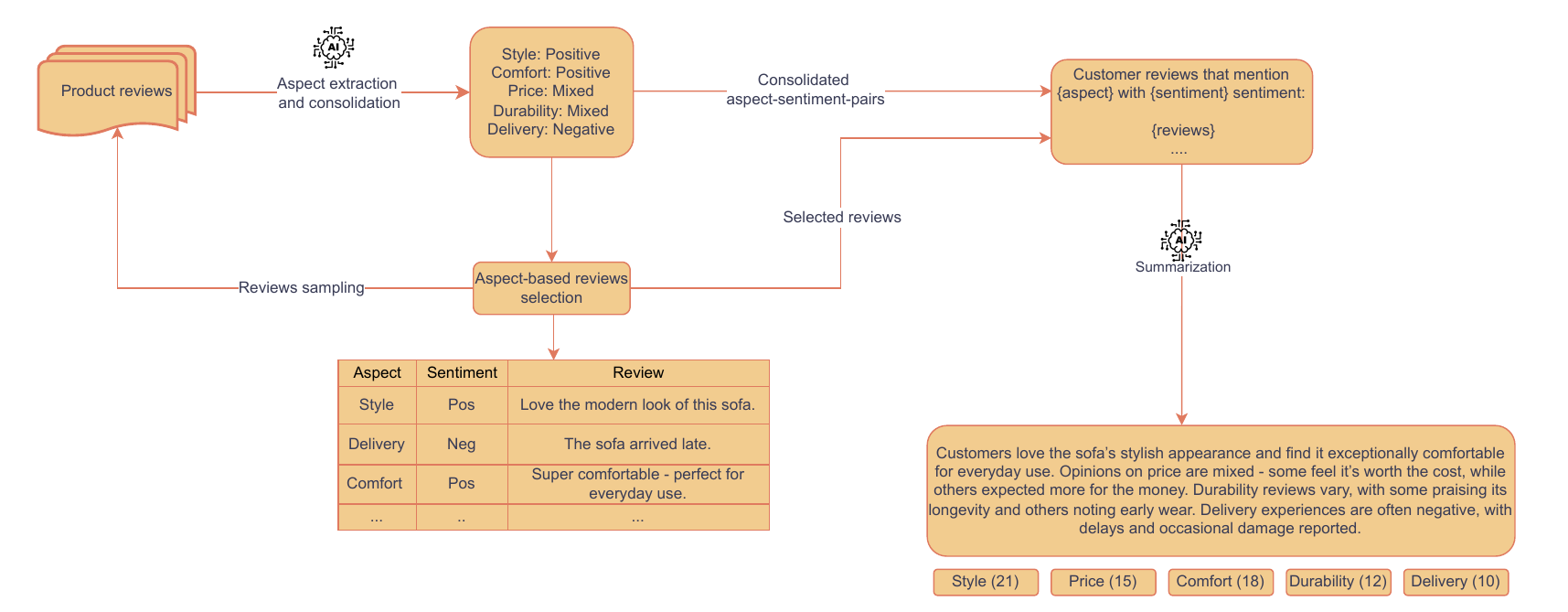}
    \caption{The figure shows an example output generated by the system. It illustrates the aspect-guided review summarization pipeline, which extracts and consolidates aspect–sentiment pairs from customer reviews to identify the top 5 most frequent aspects. The system then generates a product summary using prompts built from representative reviews for each selected aspect-sentiment pair. The output includes selected aspects with counts and supporting reviews, along with a summary grounded in them.}
    \label{fig:pipeline}
\end{figure*}

\section{Introduction}

In e-commerce platforms, customer reviews offer valuable insights into product quality and user experience. However, the overwhelming volume and repetitive nature of these reviews often make it difficult for customers to quickly find the information most relevant to their needs. To address this, platforms can leverage LLMs to automatically generate concise summaries of review content, helping customers more easily discover relevant product information. At the same time, LLMs are prone to hallucination, omission of important facts, and misrepresentation of products, particularly when summarizing large and noisy customer reviews. In addition, review sets for popular products can exceed the effective context window of current models, further increasing the risk of factual errors, as demonstrated in multiple long-context summarization applications \citep{kim2024fablesevaluatingfaithfulnesscontent, bai-etal-2024-longbench, liu-etal-2024-lost}.

Previous work has proposed methods to mitigate these issues. \citet{brazinskas-etal-2021-learning} jointly train a review selector and summarizer, enabling the system to focus on the most informative content and produce more accurate summaries. \citet{soltan-etal-2022-hybrid} propose a hybrid approach that combines extractive and abstractive summarization: they first apply Latent Semantic Analysis \citep{lsa} to select the top-$k$ most informative sentences from reviews, and then use these as input to train an abstractive summarizer. However, the creation of high-quality labeled datasets is costly, and retraining models to handle new or evolving review content adds operational overhead in production environments.

Aspect-guided review summarization offers a more controllable alternative by building on ABSA, a well-established task focused on identifying specific aspects mentioned in text and determining the sentiment expressed toward each \citep{pontiki-etal-2014-semeval}. Earlier work on ABSA relied on rule-based approaches that leveraged linguistic patterns to identify aspect terms, often by extracting prominent nouns from the text \citep{pavlopoulos-androutsopoulos-2014-aspect}. Another early direction involved the use of topic models to uncover latent aspects across reviews \citep{lakkaraju2011coherence}. With the rise of deep learning, neural network-based methods gained popularity, including approaches based on recurrent neural networks \citep{tang-etal-2016-effective, chen-etal-2017-recurrent} and convolutional neural networks \citep{xue-li-2018-aspect}. With advances in LLMs and their in-context learning capabilities \citep{brown2020language}, ABSA can now be formulated as a language generation task \citep{hosseini-asl-etal-2022-generative}, enabling scalable aspect extraction that can guide and structure the summarization process. \cite{strum} utilize aspects to construct structured summaries, but their approach does not incorporate sentiment information, which limits the controllability and nuance of the generated summaries.

Our pipeline first extracts aspect-sentiment pairs from individual reviews to identify frequently discussed product attributes along with their sentiment polarity. We then select the most frequent aspects and sample supporting reviews to construct targeted prompts that guide the summarization model to focus on specific product dimensions. This structured approach generates concise, consistent summaries while mitigating hallucination and factual inconsistency issues. Importantly, the pipeline is model-agnostic and can be applied with any pretrained LLM, making it adaptable to a wide range of deployment scenarios. Our contributions are as follows:

\begin{itemize}
    \item A production system architecture for real-time aspect-guided review summarization.
    \item Large-scale A/B test results demonstrating real-world impact on customer experiences.
    \item An open-sourced dataset with 92,000 products with total 11{,}8 million reviews, annotated aspects, and generated product summaries.
\end{itemize}

\begin{figure*}[t]
    \centering
    \includegraphics[width=\textwidth]{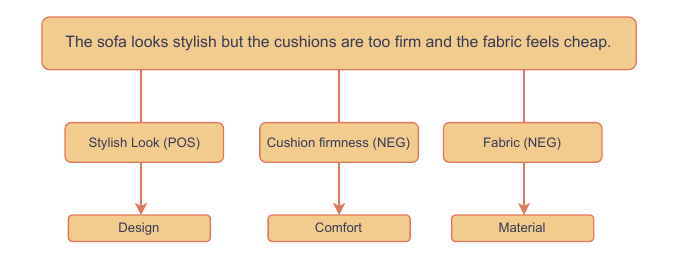}
    \caption{Illustration of the aspect extraction and consolidation process. Starting from a raw customer review, fine-grained aspect-sentiment pairs are extracted. These aspects are then mapped to broader canonical terms, resulting in a consolidated set of aspect-sentiment pairs.}
    \label{fig:aspect_transformation}
\end{figure*}


\section{Method}

Our pipeline illustrated in Figure~\ref{fig:pipeline} consists of the following stages, each described in detail in the following sections:

\begin{itemize}
  \item \textbf{Aspect extraction}: Identify product aspects and associated sentiment from individual reviews.
  \item \textbf{Aspect consolidation}: Map fine-grained or specific aspect terms to broader, high-level canonical forms.
  \item \textbf{Aspect-based review selection}: Select the most salient aspects per product and sample representative reviews mentioning each aspect.
  \item \textbf{Aspect-guided summarization}: Generate a concise product-level summary using targeted prompts constructed from the consolidated aspects and selected reviews.
\end{itemize}

We use Gemini 1.5 Flash to perform all LLM-based components of the pipeline, including aspect extraction, aspect consolidation, and final summarization. The overall architecture is modular and model-agnostic, making it straightforward to replace the LLM backbone or customize individual components as needed. 

We provide representative prompt templates used in our pipeline in Appendix. Figure~\ref{fig:aspect-extraction} shows the prompt used for extracting aspect–sentiment pairs from customer reviews. Figure~\ref{fig:aspect-consolidation} illustrates the prompt for aspect consolidation. Figure~\ref{fig:summarization} presents the prompt for generating product-level summaries.

\subsection{Aspect extraction}

The first stage of the pipeline performs aspect-level information extraction from individual customer reviews. Each review is independently processed using a pretrained LLM, prompted to extract up to five relevant product aspects and their associated sentiment labels (positive, mixed, or negative). The LLM is guided via a structured prompt and outputs a JSON object for each review.

\subsection{Aspect consolidation}
Due to the variability of natural language, the raw extracted aspects often exhibit lexical variation and differences in granularity. For example, \textit{value for money} may be mapped to the broader concept \textit{price}, \textit{assembly time} generalized to \textit{assembly}, while already high-level aspects like \textit{comfort} remain unchanged. In other cases, fine-grained terms such as \textit{shipping speed} and \textit{packaging condition} can be grouped under a shared category like \textit{delivery}.

To address this, we introduce an aspect consolidation step implemented as a second LLM-based prompting stage. We first collect all unique aspects extracted from an initial batch of data. To preserve interpretability of common terms, we compute the 95th percentile of aspect frequency - which in our case corresponds to 30 occurrences - and keep all aspects with frequencies above this threshold unchanged. Aspects with frequencies below this cutoff are consolidated by mapping them to broader canonical forms. This approach ensures that only less frequent, potentially noisy or overly specific aspects are merged, while well-established aspects remain intact. The model is prompted to map each of these to a broader or more canonical form, without relying on a fixed ontology. The resulting mappings are stored and applied across all reviews and products in the batch.

When processing new data, we reuse the cached mappings: if an aspect has already been seen, we apply its existing canonical form; otherwise, the new aspect is added to the mapping through the same consolidation process. This caching mechanism ensures consistent vocabulary over time while allowing for efficient and scalable updates as new review data becomes available. An example of this transformation is shown in the Figure~\ref{fig:aspect_transformation}.

\subsection{Aspect-based review selection}
To prepare inputs for aspect-guided summarization, we first identify the top five most frequently mentioned aspects for each product after consolidation. For each selected aspect–sentiment pair, we sample a subset of reviews in which the aspect is mentioned with the corresponding sentiment label. To maintain a reasonable context length for the summarization model, we cap the total number of input reviews at 200 per product - a hyperparameter chosen to balance coverage and efficiency. If a product has fewer than 200 reviews, all are included; otherwise, we perform random weighted sampling, drawing reviews proportionally to the frequency of each aspect–sentiment pair. This helps preserve the relative prominence of different opinions in the input to the summarization model. For algorithmic flexibility, our framework supports business criteria-based sampling such as prioritizing recent reviews or verified purchaser feedback, allowing adaptation to specific application requirements.

\subsection{Aspect-Guided Summarization}
The final stage of the pipeline generates the product-level summary based on the set of aspects and the selected supporting reviews. We frame the summarization task as an instance of aspect-guided multi-review summarization, where the model is instructed to produce a coherent, concise summary that covers the most frequent product aspects and accurately reflects the underlying customer feedback. The resulting summary is a natural language text of approximately 300–500 characters in length.

\section{Evaluation}
We evaluate our aspect-guided review summarization pipeline both offline and online to assess its quality and real-world impact. Offline, we conduct a manual review of generated summaries to measure their factual consistency and alignment with extracted aspects. Online, we deploy the system in a large-scale randomized A/B test on a live e-commerce platform, measuring its effect on key customer engagement and experience metrics.

This combination of offline and online evaluation allows us to assess both the linguistic quality of the generated summaries and their practical value to end users.

\subsection{Offline evaluation}
Our offline evaluation serves as a quality validation checkpoint before production deployment. The evaluation dataset comprises 341 products with at least 10 customer reviews each, totaling approximately 50,000 reviews. The products were strategically sampled to ensure representativeness in multiple dimensions: product categories, bestsellers, high customer engagement items, and products with varying sentiment distributions (positive-only, negative-only, and mixed reviews).

We generate aspect-guided summaries for all 341 products and perform a manual evaluation against the original review sets using the following error taxonomy:
\begin{itemize}
    \item \textbf{No errors}: The summary accurately represents all aspects with the correct sentiment.
    \item \textbf{Exaggeration / Understatement}: The summary misrepresents the overall sentiment of the customer about the product.
    \item \textbf{Minor Misrepresentation}: The summary inaccurately describes exactly one aspect of the product.
    \item \textbf{Major Misrepresentation}: The summary inaccurately describes more than one aspect of the product.
    \item \textbf{Minor Omission}: The summary fails to mention exactly one aspect of the product.
    \item \textbf{Major Omission}: The summary fails to mention more than one aspect of the product.
\end{itemize}

We initially conducted triple annotation on (10\%) of the dataset with three trained domain experts to establish reliability and identify potential subjectivity issues. Intercoder agreement measured via majority vote (at least 2 out of 3 annotators agreeing) was 70\%, with most disagreements concerning whether an error should be classified as major or minor. Following this phase, we conducted consensus discussions to refine our error taxonomy definitions and develop clearer annotation guidelines that addressed the main sources of subjectivity. The remaining 90\% of products were then independently annotated by single trained evaluators using the refined guidelines.

The evaluation revealed strong performance across quality dimensions. Of 341 summaries, 285 (84\%) had no errors. Minor problems were found in 33 summaries (11\%), including 12 cases of minor misrepresentation and 21 cases of minor omission. The main issues were identified in 15 summaries (5\%), consisting of 5 cases of exaggeration or understatement, 9 cases of major misrepresentation, and 6 cases of major omission. A few examples of evaluation can be found in the table~\ref{tab:product_analysis} of the Appendix. The input reviews are omitted due to their length; only the model-generated outputs are shown to illustrate the reasoning behind the human judgments.

\subsection{Online evaluation}
We conduct an A/B test on a large e-commerce platform comparing two versions of the product page. The control version shows the usual customer reviews without any additional features. The treatment version adds an aspect-guided summary above the reviews, along with the top n most frequently mentioned aspects. Each aspect is clickable, allowing users to filter reviews related to that specific aspect.

The primary hypothesis was that providing summaries would improve the visit-level Add to Cart Rate (ATCR) by increasing user confidence through easier access to relevant customer feedback. Secondary key performance indicators (KPIs) included the visit-level conversion rate (CVR) and the customer-level bounce rate.\footnote{Lower bounce rate indicates better user engagement.} The experiment also monitored potential negative impacts on session-level gross revenue (GRS) and page speed index to ensure that overall site performance was not degraded.

The A/B test was conducted over a three-week period in March 2025, spanning 493{,}208 products across 2,329 product classes, where each product class is a large categorical grouping such as wall art, area rugs, or coffee accessories. The experiment was carried out on English-language customer reviews. All reported metrics were statistically significant with \( p = 0.10 \). The ATCR increased by 0.3\%. Secondary metrics also improved: CVR increased by 0.5\%, and the bounce rate at the customer level decreased by 0.13\%. No statistically significant negative effects were observed on GRS or the page speed index.

\section{Real-Time System Deployment}
The pipeline is deployed in a real-time production environment to automatically generate and maintain up-to-date product review summaries. For newly listed products, a pipeline is triggered once the product accumulates at least 10 customer reviews, ensuring sufficient input for meaningful aspects extraction and summarization. For products that already have a summary, the system monitors review growth and re-triggers the pipeline whenever the number of new reviews reaches 10\% of the existing review count. This threshold-based refresh mechanism allows the summaries to dynamically adapt to evolving customer feedback without requiring manual intervention. Scalability is achieved in two ways: (1) by reusing cached aspect mappings, which eliminates the need to run the aspect consolidation step for previously seen aspects; and (2) by sampling reviews, which limits the input context length for the summarizer LLM.

\section{Dataset}

To support further research in aspect-guided review summarization, we are open-sourcing a subset of the anonymized production data used in our A/B test. The dataset covers 92{,}000 products from the 1000 most frequent product classes as observed in our A/B test sample. Each product includes between 100 and 300 reviews, resulting in 11{,}8 million reviews in total. The dataset features an average review length of 124 characters and an average of 129 reviews per product. Unlike other open-source datasets that provide customer reviews e.g., \cite{hou2024bridging}, ours pairs reviews with high-quality, production-level summaries, making it particularly valuable for training and evaluating summarization models in real-world settings.

The dataset released includes two tables. The customer reviews table contains review\_id, product\_id, review\_text, and a JSON field with extracted aspect-sentiment pairs. Before aspect consolidation, the data contained 178,054 distinct aspects. After consolidation, this number was reduced to 19,014, significantly improving consistency of the aspects extracted between products. Only the consolidated aspects are included in the data release. The most frequent are presented in Table~\ref{tab:top_aspect_stats}.
The product summaries table contains product\_id, product\_class and the corresponding natural language summary generated by our pipeline. 

\begin{table}[ht]
\centering
\small
\begin{tabular}{lrrrr}
\toprule
\textbf{Aspect} & \textbf{Count} & \textbf{Pos.} & \textbf{Neg.} & \textbf{Mix.} \\
\midrule
Quality     & 2,751,718 & 90.24 & 8.43 & 1.33 \\
Assembly    & 2,580,429 & 76.35 & 17.28 & 6.37 \\
Appearance  & 2,524,263 & 96.49 & 2.46 & 1.05 \\
Color       & 1,637,444 & 78.07 & 12.85 & 9.08 \\
Size        & 1,624,931 & 74.73 & 16.12 & 9.15 \\
Style       & 1,566,926 & 98.18 & 0.70 & 1.12 \\
Comfort     & 1,561,803 & 87.17 & 8.42 & 4.41 \\
Aesthetics  & 1,550,268 & 98.84 & 0.68 & 0.48 \\
Sturdiness  & 1,530,147 & 91.61 & 7.13 & 1.26 \\
Value       & 1,477,811 & 89.05 & 8.45 & 2.50 \\
\bottomrule
\end{tabular}
\caption{Top 10 most frequent aspects with sentiment distribution (\%).}
\label{tab:top_aspect_stats}
\end{table}

\section{Limitations}

Although effective, the performance of the pipeline depends on the quality of its individual components, including aspect extraction, consolidation, review selection and summarization. Inconsistent or noisy outputs at any stage can degrade the overall quality of the summaries. Crucially, the choice of the underlying LLM also plays a central role, as it drives all LLM-based stages and directly impacts accuracy, fluency, and alignment with customer sentiment. Performance may further decline for rare product classes with limited review coverage or highly domain-specific terminology. Future work could explore improved robustness across a broader range of domains, languages, and product categories, as well as better alignment between extracted aspects and user intent.

\section{Conclusions}

We present a production-grade end-to-end pipeline for generating product review summaries grounded in key product aspects. The modular and flexible design of the system enables experimentation with individual components and supports real-time adaptation as new reviews are added. The approach demonstrated effectiveness in an online A/B test, yielding statistically significant improvements in key customer engagement metrics. We hope that the open source dataset will support future research at the intersection of aspect-based sentiment analysis and summarization tasks.

\section*{Acknowledgments}
We thank our colleagues across multiple teams at Wayfair.  
In particular, we acknowledge the valuable contributions of John Soltis, Kyle Buday, Alsida Dizdari, Karan Manchanda, Karan Bhatia, Trevor Truog, Vipul Dalsukrai, Leo Kin, Kanika Sawhney, Kaitlyn Yan, Dan Lachapelle, and Nick Coleman.

\bibliography{custom}

\clearpage
\appendix

\section*{Appendix A. Prompt Templates}
\label{sec:appendix-prompts}

\begin{figure}[ht]
  \centering
  \includegraphics[width=\columnwidth]{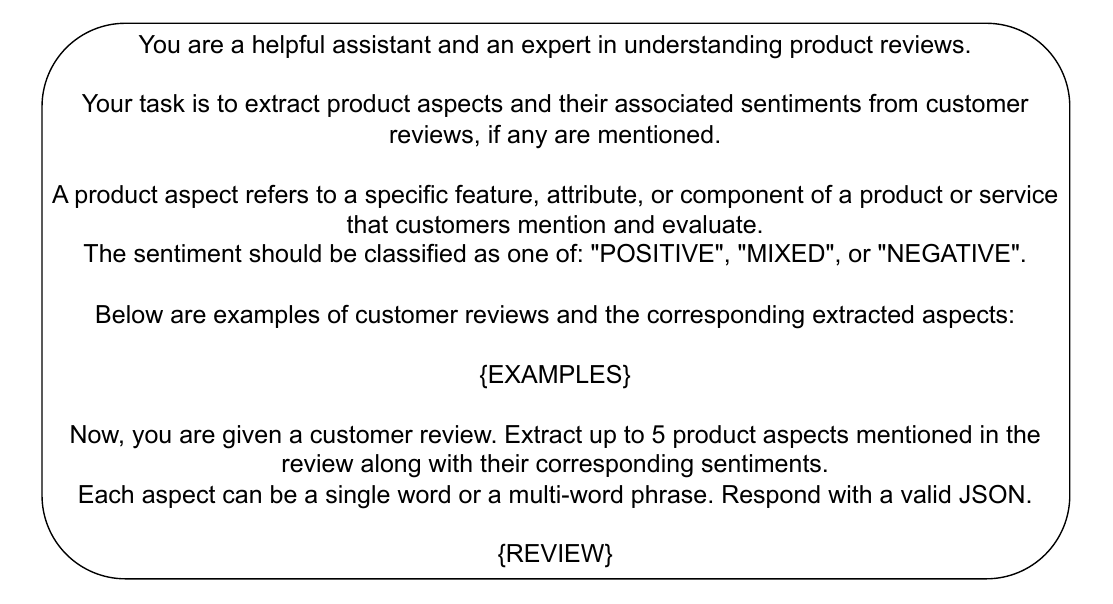}
  \caption{Prompt template for aspect extraction.}
  \label{fig:aspect-extraction}
\end{figure}

\begin{figure}[ht]
  \centering
  \includegraphics[width=\columnwidth]{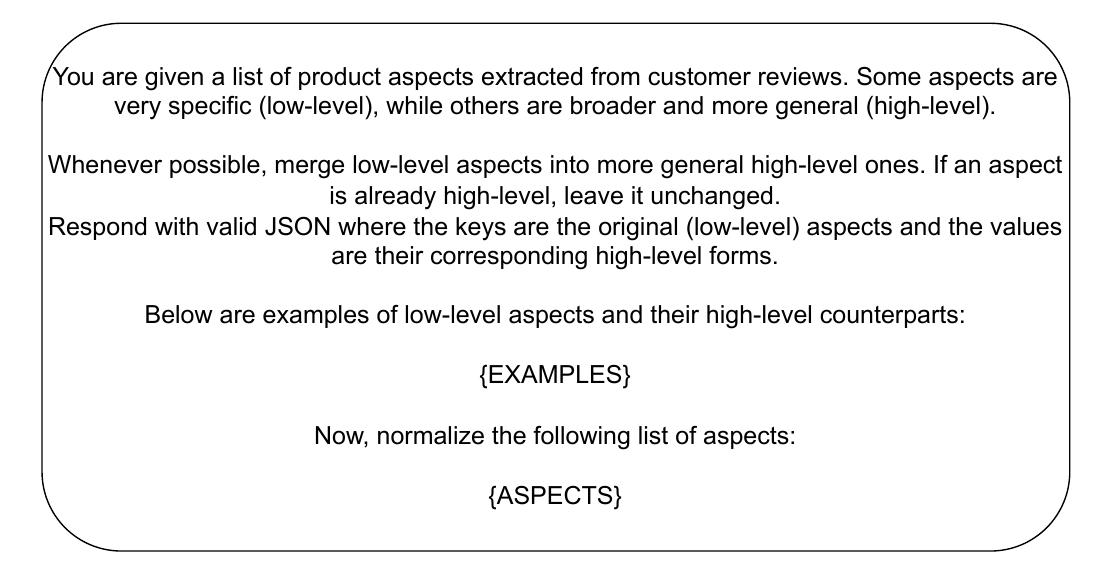}
  \caption{Prompt template for aspect consolidation.}
  \label{fig:aspect-consolidation}
\end{figure}

\begin{figure}[ht]
  \centering
  \includegraphics[width=\columnwidth]{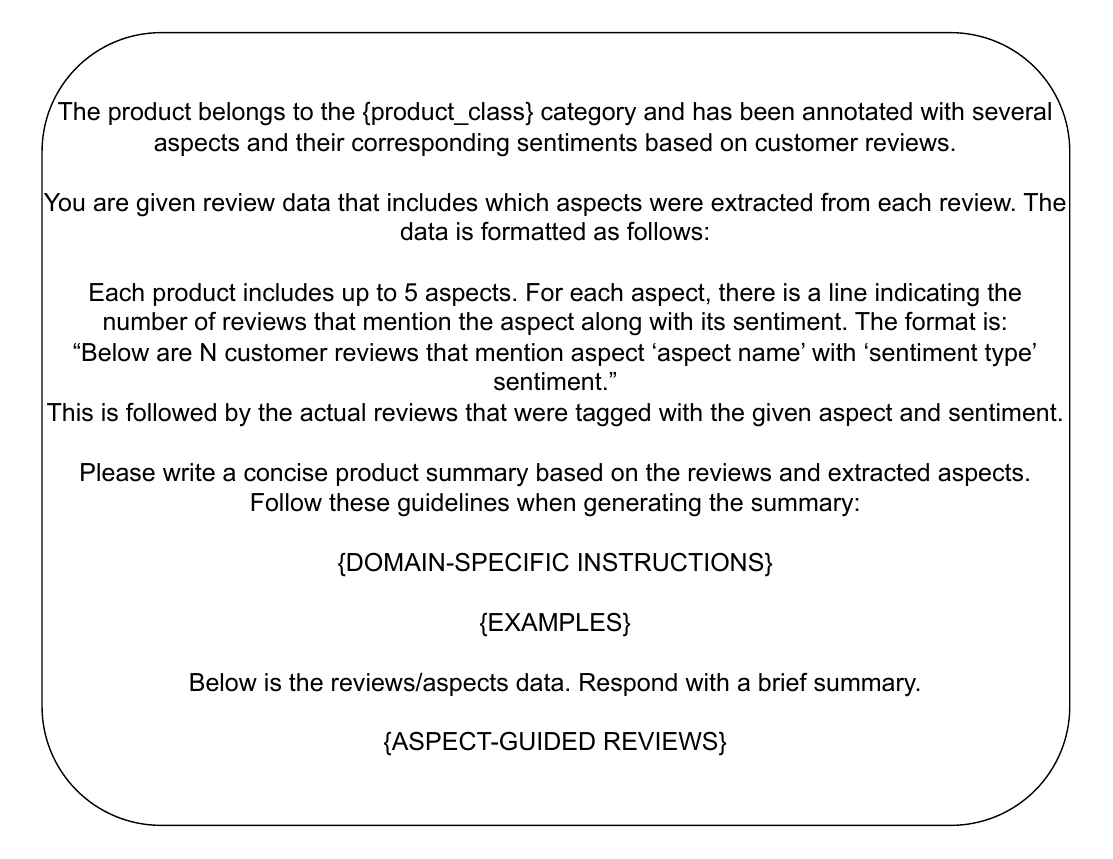}
  \caption{Prompt template for summarization.}
  \label{fig:summarization}
\end{figure}

\clearpage

\section*{Appendix B. Evaluation Examples}
\label{app:eval-examples}

\begin{table}[htbp] 
\centering 
\begin{tabular}{|p{4.5cm}|p{3.5cm}|p{2.5cm}|p{3.5cm}|} 
\hline 
\textbf{Summary} & \textbf{Aspects} & \textbf{Decision} & \textbf{Reason} \\ 
\hline 
Customers praise this product for its excellent value and supportive firmness, finding it comfortable and pain-relieving. Many appreciate its thickness and stylish design. However, some experienced inconsistencies in firmness, with some reporting it softer than expected, and others noting that the mattress didn't fully expand to its advertised thickness. & Value: POS\newline Thickness: NEG\newline Support: POS\newline Style: POS\newline Softness: MIX & Minor misrepresentation & 35 reviews stating the mattress is firm and one stating it is softer than expected. The softness aspect is misrepresented. \\ 
\hline 
Customers love the product's softness and thickness, praising its luxurious feel and substantial weight. The style is also highly regarded, complementing various decor styles. Many find it to be good value for the price. However, some note that the rug sheds excessively and its durability is questionable after washing. & Value: POS\newline Texture: POS\newline Style: POS\newline Softness: POS & Minor omission & Multiple customers questioned product durability after washing, but the durability aspect despite being mentioned in the summary was not extracted as one of the key aspects. \\ 
\hline 
This product offers good value and weather resistance, with customers praising its sturdiness and ability to withstand strong winds and rain. Assembly can be challenging, requiring multiple people and careful attention to instructions. Some users reported minor issues with leaks and misaligned parts. & Wind Resistance: POS\newline Weatherproof: POS\newline Value: POS & Exaggeration / Understatement & The product issues are much more major than just leaks and misaligned parts. Some of the customers mentioned how it collapses with minor snowfalls and other issues with rain / wind. \\ 
\hline 
\end{tabular}
\caption{Different types of errors during evaluation.}
\label{tab:product_analysis}
\end{table}

\end{document}